\documentclass[letterpaper]{article} 
\usepackage{aaai2026}  
\usepackage{times}  
\usepackage{helvet}  
\usepackage{courier}  
\usepackage[hyphens]{url}  
\usepackage{graphicx} 
\usepackage{makecell}
\usepackage{multirow}
\usepackage{natbib}  
\usepackage{caption} 
\usepackage{algorithm}
\usepackage{algorithmic}

\usepackage{multirow}
\usepackage{booktabs}
\usepackage{appendix}
\usepackage{amsthm}
\theoremstyle{definition} 
\newtheorem{definition}{Definition}[section]
\usepackage{amsmath}
\usepackage{amsfonts}
\usepackage{bm}
\usepackage{array}
\usepackage{subfigure}
\nocopyright

\usepackage{newfloat}
\usepackage{listings}
\DeclareCaptionStyle{ruled}{labelfont=normalfont,labelsep=colon,strut=off} 
\floatstyle{ruled}
\newfloat{listing}{tb}{lst}{}
\floatname{listing}{Listing}
\title{When Noisy Labels Meet Class Imbalance on Graphs: A Graph Augmentation Method with LLM and Pseudo Label}
\author{
    Riting Xia\textsuperscript{\rm 1},
    Rucong Wang\textsuperscript{\rm 1},
    Yulin Liu\textsuperscript{\rm 1},
    Anchen Li\textsuperscript{\rm 2},
    Xueyan Liu\textsuperscript{\rm 2},
    Yan Zhang\textsuperscript{\rm 1}
}
\affiliations{
    \textsuperscript{\rm 1}College of Computer Science, Inner Mongolia University, Hohhot, 010021, China\\

    \textsuperscript{\rm 2}School of Computer Science and Technology, Jilin University, Changchun, Jilin, 130012, China\\
    xiart19@mails.jlu.edu.cn,
    wrc@mail.imu.edu.cn,
    cslyl@imu.edu.cn,
    liac@jlu.edu.cn,
    xueyanliu@jlu.edu.cn,
    yanz19@mails.jlu.edu.cn
  
}

\usepackage{bibentry}

\begin{document}

\maketitle

\begin{abstract}
Class-imbalanced graph node classification is a practical yet underexplored research problem. Although recent studies have attempted to address this issue, they typically assume clean and reliable labels when processing class-imbalanced graphs. This assumption often violates the nature of real-world graphs, where labels frequently contain noise. Given this gap, this paper systematically investigates robust node classification for class-imbalanced graphs with noisy labels. We propose GraphALP, a novel \textbf{Graph} \textbf{A}ugmentation framework based on \textbf{L}arge language models (LLMs) and \textbf{P}seudo-labeling techniques. Specifically, we design an LLM-based oversampling method to generate synthetic minority nodes, producing label-accurate minority nodes to alleviate class imbalance. Based on the class-balanced graphs, we develop a dynamically weighted pseudo-labeling method to obtain high-confidence pseudo labels to reduce label noise ratio. Additionally, we implement a secondary LLM-guided oversampling mechanism to mitigate potential class distribution skew caused by pseudo labels.
Experimental results show that GraphALP achieves superior performance over state-of-the-art methods on class-imbalanced graphs with noisy labels.
\end{abstract}

\section{Introduction}\label{section1}
Graphs with nodes and their edges are ubiquitous in the real world, such as knowledge graphs~\cite{DBLP:conf/aaai/WangM0Z25}, recommendation networks~\cite{DBLP:conf/www/Li025}, and citation networks~\cite{DBLP:conf/www/ZhongL24}. 
Within the graph learning paradigm~\cite{DBLP:journals/corr/abs-2502-12412}, node classification emerges as a fundamental graph analysis task whose unique challenges have inspired numerous algorithmic advances.
However, this task commonly faces the challenge of class imbalance, where the majority nodes dominate model training, resulting in bias against the minority nodes~\cite{DBLP:journals/tkde/LiuLCWHH25}.
For example, in fake account detection, extreme class imbalance occurs where legitimate users dominate ($>99\%$) while bot accounts are severely underrepresented ($<1\%$), causing models to fail in detecting fraudulent activity. 

Existing research on addressing class imbalance graphs can be categorized into two approaches: model-based and data-based methods~\cite{DBLP:journals/chinaf/XiaZZLY24}. Model-based methods achieve class balance by algorithmically optimizing the model's focus on minority nodes. Data-based methods employ augmentation strategies to increase the number of minority nodes, thereby mitigating class imbalance in graphs.

Although significant progress has been made in class-imbalanced node classification, existing approaches rely on the unrealistic label reliability assumption. This assumption fails to account for the inherent label noise in real-world graphs, where node mislabeling occurs due to annotation errors and intrinsic ambiguity~\cite{DBLP:journals/nn/LiLGLQW25}.
Real-world scenarios often exhibit coexisting class imbalance and label noise.
For example, in citation networks, certain paper categories dominate quantitatively while simultaneously containing mislabeled nodes. When there is label noise in the graphs, the performance of existing graph learning methods significantly deteriorates, as shown in Figure \ref{Fig. 5}.
To achieve robust node classification, it is crucial to develop methods to address both class imbalance and label noise in graphs. 

This paper presents the first explicit study of this previously unexplored, yet highly pragmatic research direction.
Several challenges must be addressed to adapt the graph learning model to class-imbalanced graphs with noisy labels. \textbf{Challenge 1: the impact of class imbalance on noise processing.} Many previous works have emerged to address the problems of noisy labels on graphs~\cite{DBLP:conf/nips/WangSZWFHB24}. These methods rely on the class balance assumption. However, this assumption is invalid due to the class-imbalanced issue. For example, commonly used methods for handling noisy labels based on confidence scores cannot be directly applied, as minority nodes tend to have lower confidence scores, which may lead to misidentification of minority node labels as noisy labels, thereby causing the model to favor the majority classes. \textbf{Challenge 2: the impact of label noise on class imbalance processing.} Methods designed for class imbalance graphs primarily employ oversampling and reweighting techniques to balance classes, but do not account for the propagation of noisy labels, resulting in the accumulation of label errors. In summary, class imbalance and label noise interact synergistically, invalidating isolated or naive combined approaches. Therefore, there is an urgent need to propose a novel approach to address this problem.


Based on these observations, we propose GraphALP, a novel graph augmentation method that integrates LLM and pseudo label. GraphALP can better balance classes and reduce the impact of label noise on graph node classification through two ways. (1) 
To mitigate class imbalance and label noise, GraphALP generates synthetic minority nodes using LLM. We first generate minority node text via LLM, then learn their representations through language model, ultimately achieving balanced distribution in graphs while generating accurate labels for synthetic nodes.
(2) To mitigate label noise, GraphALP introduces a dynamically weighted pseudo-labeling method to obtain pseudo labels. Furthermore, we implement a secondary LLM-guided oversampling to mitigate class distribution skew caused by pseudo-labeling. The main contributions are as follows:

\begin{itemize} 
\item We are the first to focus on class-imbalanced graph learning with noisy labels, a challenging problem in the field. 
\item GraphALP is a novel framework based on LLM and pseudo label for class-imbalanced graphs with label noise. It synthesizes minority nodes with LLM and employs pseudo-labeling method to improve supervision. This is the first work to tackle these challenges via LLM.

\item We conduct extensive experiments on real-world datasets. The results demonstrate that our method consistently outperforms existing state-of-the-art approaches. 
\end{itemize}


\section{RELATED WORK}\label{section2}

\subsection{Class-imbalanced Graph Learning}\label{section2.1}
Class-imbalanced graph refers to a graph in which the distribution of classes is highly skewed. This skewed distribution biases the model towards the majority class during training, degrading its classification performance on minority nodes~\cite{DBLP:journals/csur/MaTMC25}. Current research can be divided into algorithm- and data-level methods~\cite{DBLP:conf/www/XuWZW0024}. 

Data-level methods~\cite{DBLP:journals/chinaf/XiaZZLY24} rely on oversampling to balance class distributions, while algorithm-level methods~\cite{DBLP:conf/iccv/TaoSYCWYDZ23} enhance models to class imbalance by modifying model architectures or optimization objectives. Among them, oversampling~\cite{DBLP:journals/tkde/AbdiH16} can generate minority nodes, effectively alleviating the imbalance. For example, GraphSMOTE \cite{DBLP:conf/wsdm/ZhaoZW21}  generates synthetic nodes by SMOTE \cite{DBLP:journals/tkde/SunLZ25}, and adds edges to these nodes by the decoder. GraphMixup \cite{DBLP:conf/pkdd/WuXGLTL22} constructs a semantic relation space through disentanglement and synthesizes minority nodes by oversampling at the semantic level. 
Graph-DAO~\cite{DBLP:journals/chinaf/XiaZZLY24} generates minority nodes before applying graph neural network.
GQEO~\cite{DBLP:journals/nn/DaiWZDC25} introduces a generalized quadrilateral element oversampling that uses k-NN to resolve imbalance. 
Although effective, these methods fail to account for the impact of label noise, particularly for minority classes, where the presence of incorrect labels leads to the generation of erroneous minority nodes.  

\subsection{Label Noise Graph Learning}\label{section2.2}
Graphs are vulnerable to label noise, which severely degrades the generalization of the model~\cite{DBLP:conf/nips/WangSZWFHB24}. 
Many approaches have been proposed to handle noisy labels on graphs, including but not limited to contrastive learning~\cite{DBLP:journals/nn/LiLLQW24}, label correction~\cite{DBLP:conf/pakdd/LiYC21}, and pseudo-labeling methods~\cite{DBLP:journals/tkde/XiaLXTWLL24, DBLP:conf/kdd/Dai0W21}. Among these methods, pseudo-labeling methods have emerged as a popular method for addressing label noise by augmenting the training set with selected high-confidence nodes. For example, GNN Clean~\cite{DBLP:journals/tkde/XiaLXTWLL24} generates pseudo labels by aggregating information from neighboring nodes, while NRGNN~\cite{DBLP:conf/kdd/Dai0W21} mines accurate pseudo labels through strategic connections between labeled and unlabeled nodes. However, existing methods fail to account for class imbalance, and their pseudo-labeling accuracy is vulnerable to noise labels. 

\subsection{Language Models on Graphs.}\label{section2.3}
Large Language Models (LLMs)~\cite{DBLP:conf/acl/LongWXZDCW24} can acquire comprehensive linguistic knowledge through large-scale corpus training. Through prompt engineering, LLMs can be effectively guided to perform knowledge generation and reasoning tasks. In recent years, LLMs are used to graphs~\cite{DBLP:conf/www/KhoshraftarAH25, DBLP:conf/aaai/WangCOWHSGXZCLZ24, wang2024large}. However, existing approaches focus on text-attributed graphs and have yet to adequately address the issue of class imbalance and label noise on graphs. We aim to leverage LLMs to establish supervisory signals for enhancing graph reliability, simultaneously mitigating both class imbalance and label noise issues, rather than directly performing data augmentation on graphs.



\begin{figure*}[ht]
\centering
\includegraphics[scale=0.6]{ 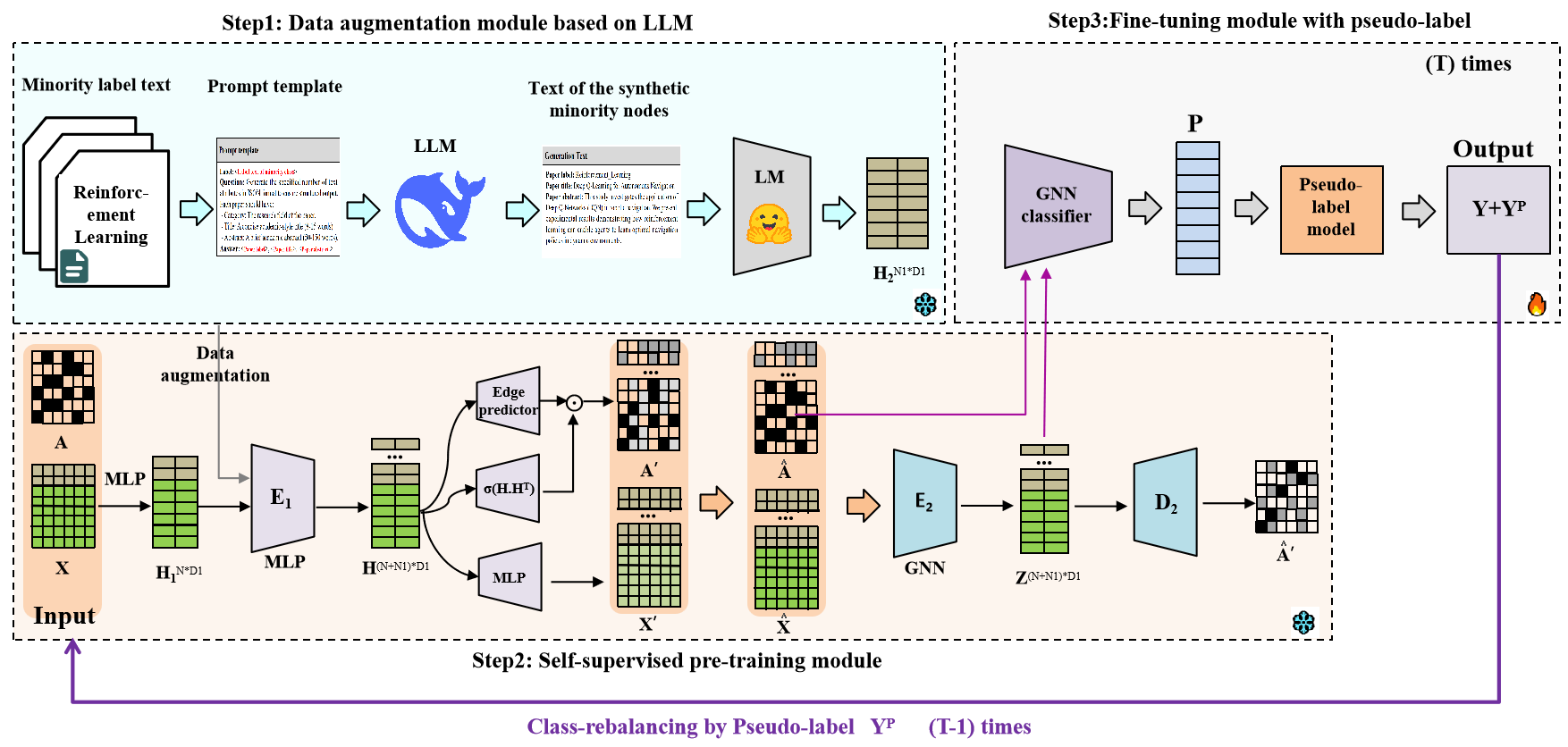}
\caption{The architecture of GraphALP.}
\label{Fig. 2}
\end{figure*}

\section{PRELIMINARIES}\label{section3}

\subsection{Problem Definition}\label{section3.1}
\begin{definition}\label{Definition 1}
(Class-imbalanced graphs with noisy labels). $\mathbf{G}  = (\textbf{V}, \textbf{X}, \textbf{E}, \textbf{C}, \textbf{Y}_{t})$ is a class-imbalanced graph with noisy labels, which contains the node set $\textbf{V}=\left \{  {v_{1}, \dots, v_{n}} \right \}$, class set $\textbf{C}= \left \{  {C_{1}, \dots, C_{k}} \right \}$, edge set $\textbf{E}=\left \{ {e_{ij} = (v_{i},v_{j}) } \right \}$, attribute set $\textbf{X}= \left \{  {x_{1}, \dots, x_{m}} \right \}$, and label set $\textbf{Y}_{t}=\left \{  {\hat{y}_{1}, \dots, \hat{y}_{s}} \right \}$, where $n$, $k$, $m$ and $s$ are the quantity of nodes, class types, attribute types and labeled nodes, respectively. $\textbf{A}\in \{0,1\} ^{n\times n}$ denote the adjacency matrix, where $a_{ij} = a_{ji} =1$ if $e_{ij} \in \mathbf{E}$, otherwise $a_{ij}= 0$. $\left | C_{k} \right | $ denotes the node number of class $k$, and $\rho = \frac{min_{k}\left | C_{k} \right | }{max_{k}\left | C_{k} \right | }$ denotes the imbalance ratio, $\rho$ is small (there exists $\left | C_{i} \right | \ll \left | C_{j} \right |$). The labels $\hat{y}_{i}$ contain noise. 
\end{definition}

\subsection{Graph Neural Networks}\label{section3.2}
Graph Neural Networks (GNNs) have emerged as the predominant methodology for graph learning~\cite{DBLP:conf/aaai/0005GLLCZ25, DBLP:conf/nips/FangZYWC23}. We introduce GNNs for node classification. A GNN comprises three functions: message passing, aggregation, and node update. Its layer-$l$ formulation is:
\begin{equation}\label{equ:1}
\begin{aligned}
\textbf{h}_{v}^{l+1} = f_{l}(\textbf{x}_{v}^{l}, \varphi_{l}(\left \{ m_{l}(\textbf{x}_{v}^{l}, \textbf{x}_{u}^{l}, e_{v,u}) | u\in N(v)\right \} ) ),
\end{aligned}
\end{equation}
where $m_{l}$, $\varphi_{l}$, and $f_{l}$ are the message passing, aggregation, and node update functions of layer $l$, respectively. $\textbf{x}_{v}^{l}$ and $\textbf{x}_{u}^{l}$ are the representations of node $v$ and $u$ at layer $l$. $N(v)$ is the adjacent nodes of node $v$, $e_{v,u}$ is the edge weight. Our goal is to learn a GNN classifier $f(\mathbf{G},\textbf{Y}_{t}) \to \textbf{Y}$ that performs effectively for both class imbalance and label noise.


\section{METHODOLOGY}\label{section4}
As shown in Figure \ref{Fig. 2}, GraphALP contains three key modules:
(1) \textbf{Data Augmentation Module based on LLM}, which generates synthetic minority node text via LLM, encodes it into representations via language model (LM), and balances graph class distribution. This module mitigates imbalance and reduces label noise with high-quality synthetic nodes.
(2) \textbf{Self-Supervised Pre-Training Module}, which constructs a class-balanced graph with synthetic nodes and learns representations in a self-supervised manner. The resulting representations are used for downstream tasks.
(3) \textbf{Fine-tuning Module with Pseudo-label}, which employs pseudo-labeling techniques to generate high-confidence labels, expanding the training set while further reducing label noise. Moreover, the LLM is used to rebalance the classes.
By integrating these modules, GraphALP achieves more robust results for class imbalance graphs with noisy labels.

\subsection{Data Augmentation Module based on LLM}\label{section4.1}

\subsubsection{Minority Node Generation by LLM}\label{section4.1.1}
Oversampling is an effective approach for class imbalance~\cite{DBLP:journals/nn/DaiWZDC25}. However, it generates synthetic nodes from existing ones, which lack semantic richness and lead to overfitting. Moreover, it can propagate label noise, particularly when mislabeled nodes produce erroneous synthetic nodes.
Based on the above analysis, we introduce LLM to generate synthetic minority nodes with minority labels to capture rich semantic information and effectively reduce the label noise ratio.

\begin{figure}[ht]
\centering
\includegraphics[scale=0.23]{ 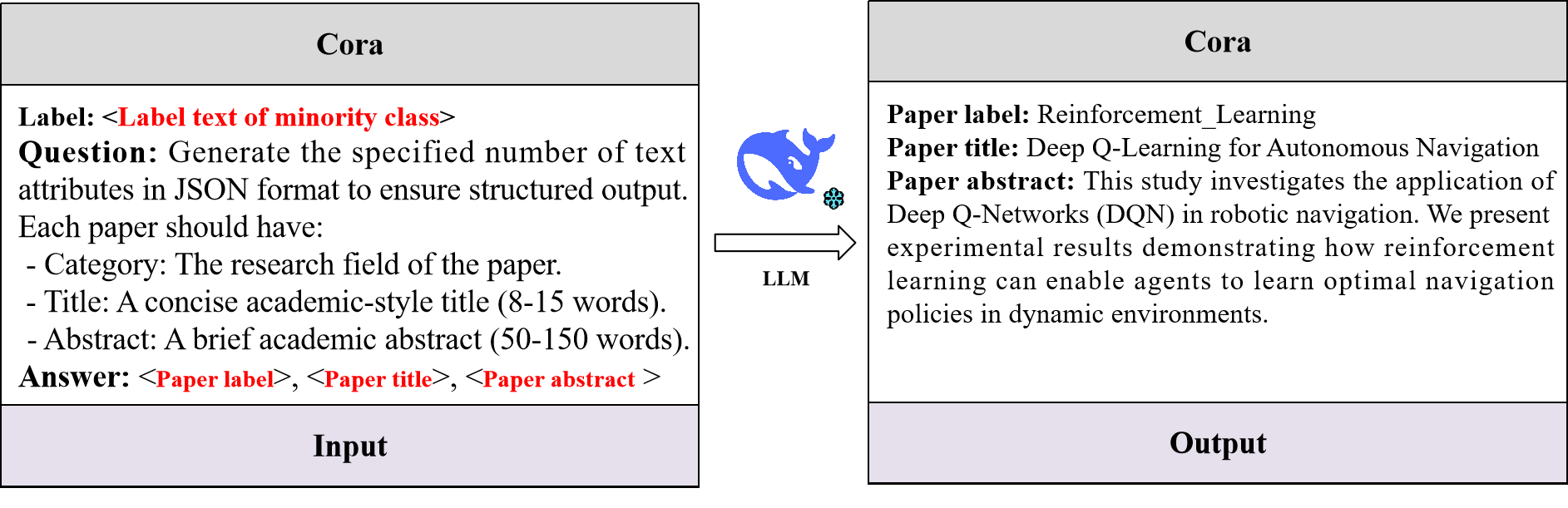}
\caption{Prompt and node text generated by LLM on Cora.}
\label{Fig. 3}
\end{figure}

Specifically, we employ a prompt $\theta_{LLM}$ to generate synthetic minority nodes. For example, for the citation networks, we generate the nodes that contain the titles and abstracts of the paper (prompt templates and LLM-generated texts are illustrated in Figure \ref{Fig. 3}). The process is as follows
\begin{equation}\label{equ:2}
\begin{aligned}
T_i^{prompt} = \theta_{LLM}(T_i^{label}),
\end{aligned}
\end{equation}
\begin{equation}\label{equ:3}
\begin{aligned}
y_i, T_{i'}^{attr} = LLM(T_i^{label}, T_i^{prompt}),
\end{aligned}
\end{equation}
where $y_i$ and $T_{i'}^{attr}$ are the label and node text of the synthetic node $v_i^{'}$, respectively. $T_i^{label}$ represents the label text of node $v_i$, and $T_i^{prompt}$ refers to the prompt template.
\subsubsection{Minority Node Representation Initialization by LM}\label{section4.1.2}
After obtaining the text of the synthetic nodes $v_i^{'}$, we encode their text using LM to obtain initial node representations. Specifically, we employ the SentenceBERT sentence embedding model jinai-embedding-v3~\cite{DBLP:journals/corr/abs-2409-10173} to extract initial node representations $\bm{h}_{v_i^{'}}$ from node text. 
\begin{equation}\label{equ:4}
\begin{aligned}
\bm{h}_{v_i^{'}} = LM(T_{i'}^{attr}).
\end{aligned}
\end{equation}

\subsection{Self-Supervised Pre-Training Module}\label{section4.2}
GNN is an effective method for graph learning. However, their neighbor aggregation strategy often biases minority classes toward majority classes. Although some methods use oversampling to balance classes before GNN, their synthetic nodes from existing nodes are susceptible to overfitting~\cite{DBLP:journals/chinaf/XiaZZLY24}. To address this issue, we propose self-supervised pre-training module, which contains: (1) AE based module. Augment the original graph with LLM-generated synthetic nodes to build a class-balanced graph before performing neighborhood aggregation in the GNN. (2) GAE based module. Node representations are learned through a graph autoencoder (GAE).

\subsubsection{AE based Module}\label{section4.2.1}
We first feed the node attribute vectors $\bm{X}$ from the original graph into a Multi-Layer Perceptron (MLP) to learn latent representations:
\begin{equation}\label{equ:5}
\begin{aligned}
\bm{H}_1 = MLP_{1}(\bm{X}).
\end{aligned}
\end{equation}

Then, synthetic minority nodes are generated using the LLM to balance the classes. The synthetic minority node representations are obtained by
\begin{equation}\label{equ:6}
\begin{aligned}
\bm{H}_2 = LM(LLM(T_i^{label}, T_i^{prompt})).
\end{aligned}
\end{equation}

To align $\bm{H}_1$ with $\bm{H}_2$, we use an MLP encoder to project them into a common latent space. 
\begin{equation}\label{equ:7}
\begin{aligned}
\bm{H}_1 = MLP_{2}(\bm{H}_1), 
\bm{H}_2 = MLP_{2}(\bm{H}_2),
\end{aligned}
\end{equation}
where the final node representation matrix is $\bm{H}=Concat(\bm{H}_1,\bm{H}_2)$, $Concat(.)$ is the concatenation function. 

The decoder contains attribute reconstruction, edge predictor, and structural reconstruction decoders. For attribute reconstruction decoder, we use MLP function. 
\begin{equation}\label{equ:8}
\begin{aligned}
\pmb{X}^{'} = MLP_3 (Concat(\bm{H}_1, \bm{H}_2)).
\end{aligned}
\end{equation}

For structural reconstruction decoder, we use inner-product to reconstruct graph structure.
\begin{equation}\label{equ:9}
\begin{aligned}
\pmb{A}_1 = \sigma  (\bm{H}\cdot \bm{H}^{T}).
\end{aligned}
\end{equation}

To learn more discriminative representations, we regularize the autoencoder with both attribute and structural reconstruction losses, formulated as:
\begin{equation}\label{equ:10}
\begin{aligned}
\mathcal{L}_{X} = \left \| \bm{X}^{'}-\bm{X} \right \|_{F}^{2}, \mathcal{L}_{A} = \left \| \bm{A}_1-\bm{A} \right \|_{F}^{2}.
\end{aligned}
\end{equation}

Although the structural reconstruction decoder can generate edges for synthetic minority nodes, connections derived solely from the inner product function remain unreliable. Therefore, we introduce an edge predictor to constrain the model to obtain more accurate edges.
Specifically, we establish initial edges through cosine similarity $sim(\cdot)$ between newly generated nodes $g_i$ and original nodes $v_j$.
\begin{equation}\label{equ:11}
\begin{aligned}
e(\bm{h}_{g_{i}}, \bm{h}_{v_{j}})=\left\{\begin{matrix}{1 , if \begin{matrix}
  &  
\end{matrix}sim(\bm{h}_{g_{i}}, \bm{h}_{v_{j}})> \tau }
  & \\{0 , if \begin{matrix}
  &  
\end{matrix}sim(\bm{h}_{g_{i}}, \bm{h}_{v_{j}})\le  \tau }
  &
\end{matrix}\right..
\end{aligned}
\end{equation}

In addition, to ensure topological consistency between newly added edges and the original graph, we formulate an edge prediction task where the existing edges serve as supervision signals to predict edges for synthetic nodes.
\begin{equation}\label{equ:12}
\begin{aligned}
e_1(\bm{h}_{g_{i}}, h_{\bm{v}_{j}})=MLP(\bm{h}_{g_{i}}\left |  \right | \bm{h}_{v_{j}}),
\end{aligned}
\end{equation}
\begin{equation}\label{equ:13}
\mathcal{L}_{E} = -\sum_{}^{}e_1(\bm{h}_{g_{i}}, \bm{h}_{v_{j}})log y(\bm{h}_{g_{i}}, \bm{h}_{v_{j}}),   
\end{equation}
where $\bm{e}_1(\bm{h}_{g_{i}}, \bm{h}_{v_{j}}) \in e(\bm{h}_{g_{i}}, \bm{h}_{v_{j}})$. $y(\bm{h}_{g_{i}}, \bm{h}_{v_{j}}))=1$, if have edges, else $y(\bm{h}_{g_{i}}, \bm{h}_{v_{j}}))=0$. The edges obtained by the edge predictor decoder is $\bm{A}_2$.

Finally, we compute the Hadamard product between the edge matrix $\bm{A}_2$ generated by the edge predictor and the edge matrix $\bm{A}_1$ reconstructed by the structural reconstruction decoder to obtain the final edges for the synthetic nodes.
\begin{equation}\label{equ:14}
\bm{A}^{'}=\bm{A}_1 \odot  \bm{A}_2.   
\end{equation}

\subsubsection{GAE based Module}\label{section4.2.2}
To obtain more semantic information, we introduce GAE to aggregate neighborhood information. We integrate the attributes of synthetic nodes and their edges into the original graph $G$ to construct a new class-balanced graph $\hat{G}(\hat{\textbf{X}}, \hat{\textbf{A}} )$. Then feed $\hat{G}$ to the GNN encoder to obtain node representations:
\begin{equation}\label{equ:15}
\begin{aligned}
\pmb{Z}  = \rm{GraphSage} (\hat{\textbf{X}} , \hat{\textbf{A}} ),
\end{aligned}
\end{equation}
where $\rm GraphSage(\cdot)$ is a single layer of GraphSage, the message
passing and aggregation process is:
\begin{equation}\label{equ:16}
\begin{aligned}
\pmb{h}_{g_{v}}=ReLU (\textbf{W}\cdot Concat(\hat{\textbf{X}}[v,:], \hat{\textbf{X}}\cdot \hat{\textbf{A}}[:, v])),
\end{aligned}
\end{equation} 
where $\hat{\textbf{X}}[v,:]$ is the feature of node $v$. $\hat{\textbf{A}}[:, v]$ is the $v$-$th$ column of $\hat{\textbf{A}}$, $\textbf{W}$ is the weights, and $\pmb{h}_{g_{v}}$ is the representation. 

Furthermore, we utilize a structural decoder to constrain node representations. The loss function is
\begin{equation}\label{equ:19}
\begin{aligned}
\mathcal{L}_\mathrm{{\hat{A}}}=\left \| \hat{\textbf{A}{'}}-\bm{A} \right \|_{F}^{2},
\end{aligned}
\end{equation}
where $\hat{\textbf{A}}{'}$ is reconstructed by
$\hat{\textbf{A}{'}} = \sigma  (\bm{Z}\cdot \bm{Z}^{T})$.
 
\begin{table*}[ht]
\caption{ACC, G-mean and F1 values for node classification.}
\centering
\setlength{\tabcolsep}{5pt}
\begin{tabular}{ccccccccccccc}
\toprule
                          & \multicolumn{3}{c}{Cora}                                                                   & \multicolumn{3}{c}{CiteSeer}                                                              & \multicolumn{3}{c}{Pubmed}                                                                 & \multicolumn{3}{c}{Wiki-CS}                                                                \\ \cline{2-13} 
\multirow{-2}{*}{Methods} & ACC                          & F1                           & G-mean                       & ACC                          & F1                          & G-mean                       & ACC                          & F1                           & G-mean                       & ACC                          & F1                           & G-mean                       \\ \hline
Vanilla                   & 51.17                       & 50.02                       & 47.22                       & 38.79                       & 37.37                      & 34.08                       & 60.61                       & 59.51                       & 57.83                       & {\underline {77.81}}                 & {\underline {73.43}}                 & 68.24                       \\ \hline
Oversampling              & 51.32                       & 50.50                       & 48.07                       & 39.21                       & 38.31                      & 35.54                       & 61.45                       & 60.37                       & 58.78                       & 76.45                       & 71.24                       & 67.65                       \\ \hline
Re-weight                 & 50.96                       & 50.08                       & 47.53                       & 0.3921                       &38.22                      & 35.50                       & 61.21                       & 60.20                       & 58.71                       & 76.35                       & 71.29                       & 68.11                       \\ \hline
SMOTE                    & 58.29 & 57.66 & 55.05 & 46.00 & 45.95 & 43.59 & 64.00 & 63.49 & 62.62 & 76.37 & 71.36 & 68.69                       \\ \hline
Embed-SMOTE    & 46.81 & 46.32 & 44.03 & 38.06 & 37.85 & 35.79 & 59.03 & 58.35 & 57.15 & 76.62 & 71.70 & 67.91                       \\ \hline
GraphMixup & 67.48 & 66.44 & 63.55 & 51.58 & 51.61 & 48.54 & 66.30 & 65.52 & \underline{64.02} & 75.86 & 72.39 & \underline{70.03}                 \\ \hline
GraphDAO                       & \underline{70.60} & \underline{70.33} & \underline{66.85} & \underline{57.58} & \underline{57.18} & \underline{54.48} & \underline{66.30} & \underline{66.00} & 63.90 & 77.67 & 72.86 & 66.82\\ \hline
GraphALP & \textbf{75.95} & \textbf{76.11} & \textbf{74.88} & \textbf{60.42} & \textbf{59.98} & \textbf{55.67} & \textbf{70.18} & \textbf{70.02} & \textbf{69.56} & \textbf{80.28} & \textbf{77.36} & \textbf{76.06}            
\\ \hline
Optimality gap            & {\color[HTML]{0000C7} +5.35} & {\color[HTML]{0000C7} +4.78} & {\color[HTML]{0000C7} +8.03} & {\color[HTML]{0000C7} +2.84} & {\color[HTML]{0000C7} +2.8} & {\color[HTML]{0000C7} +1.19} & {\color[HTML]{0000C7} +3.88} & {\color[HTML]{0000C7} +4.02} & {\color[HTML]{0000C7} +5.54} & {\color[HTML]{0000C7} +2.47} & {\color[HTML]{0000C7} +3.93} & {\color[HTML]{0000C7} +6.03} \\ 
\bottomrule
\end{tabular}
\label{table 2}
\end{table*}

\subsection{Fine-tuning Module with 
Pseudo-label}\label{section4.3} 
Recent studies have shown that more accurate pseudo labels can effectively enhance GNNs under noisy and limited label conditions \cite{DBLP:conf/nips/WangSZWFHB24}. However, existing methods fail to account for the class imbalance. In practice, the generated pseudo labels tend to be biased towards the majority classes (as shown in \ref{Fig. 6}), which may exacerbate the imbalance. To handle these issues, we adopt a rebalancing-enhanced pseudo-labeling method. 

We first use a GNN classifier as a pseudo-label generator.
\begin{equation}\label{equ:20}
\hat{\textbf{Y}}_P = GraphSage(\textbf{Z},\hat{\textbf{A}}).
\end{equation}

To focus on minority nodes and enhance the accuracy of pseudo-label selection, we employ an adjustable weighted cross-entropy loss for the node classification task.
\begin{equation}\label{equ:21}
\mathcal{L}_{C} = -\sum_{i=1}^{C}w_{C_{i}}\cdot y_{i}log \hat{y} _{i}, 
\end{equation}
where $w_{C_{i}}=max\left ( 1, \frac{N}{n_{i} }  \right ) $ is the weight for the $i$-$th$ class, which adjusts the contribution of each class. A larger $w_{C_{i}}$ indicates that the corresponding class is a minority class. $n_i$ is the node number of class $i$, $N$ is the total number of nodes. $y_{i}$ and $\hat{y} _{i}$ are the ground-truth label and predicted probability of 
$i$-$th$ class, respectively. $C$ is the class number.

We use a confidence-based strategy for pseudo-label assignment. Pseudo labels with higher confidence scores exhibit greater prediction reliability. Let $\hat{y}_{ic}$ denote the predicted probability of node $v_i$ belonging to class $c$. 
\begin{equation}\label{equ:22}
\textbf{Y}_P = \left \{ { \hat{y}_{ic} \in \hat{\textbf{Y}}_U \mid \hat{y}_{ic} > \tau} \right \}, 
\end{equation}
where $\hat{\textbf{Y}}_U$ is the pseudo-label predictions for the unlabeled nodes. $\tau$ is the confidence threshold. By incorporating pseudo-label, our method mitigates label noise while generating more reliable pseudo labels. Furthermore, through adaptive weighting, our method emphasizes minority nodes, thereby alleviating the model bias towards majority nodes.

To mitigate the potential exacerbation of class imbalance induced by pseudo-labeling, we integrate pseudo labels into the training set and through the self-supervised pre-training module to learn node representations. Finally, we use the GNN classifier to obtain the final node classification results.

\subsection{Loss Function}\label{section4.4}
The loss function comprises edge predictor loss
$\mathcal{L}_{E}$, pre-training reconstruction losses $\mathcal{L}_{\text{A}}$, $\mathcal{L}_{\hat{A}}$ and $\mathcal{L}_{\text{X}}$, and node classification loss $\mathcal{L}_{C}$.
During the pre-training phase, to constrain the model to learn more balanced and semantically richer representations, we adopt reconstruction losses:
\begin{equation}\label{equ:24}
\mathcal{L}_{\text{recon}} = \alpha 
\mathcal{L}_{\text{A}} + \beta \mathcal{L}_{\text{X}} + \gamma  \mathcal{L}_{\hat{A}}.
\end{equation}

For fine-tuning, we use class-balanced cross-entropy loss:
\begin{equation}\label{equ:25}
\mathcal{L}_{C} = -\sum_{i=1}^{C}w_{C_{i}}\cdot y_{i}log \hat{y} _{i} -\sum_{i=1}^{C}w_{C_{i}}\cdot y^{P}_{i}log \hat{y} _{i},  
\end{equation}
where $y^{P}_{i} \in \textbf{Y}^{P}$ is the pseudo label. The joint optimization loss is then formulated as
\begin{equation}\label{equ:26}
\mathcal{L} = \mathcal{L}_{E}+ \mathcal{L}_\text{recon} +  \mathcal{L}_{\text{C}}
\end{equation}
Through the losses of the two stages, the influence of label noise and class imbalance is gradually reduced. 

\begin{table}
\centering
\renewcommand{\arraystretch}{0.9}{
\setlength{\tabcolsep}{2mm}{
\caption{Summary statistics of graph datasets.}
\label{table 1}
\begin{tabular}{cccccc} 
\toprule
Datasets    & Classes & Nodes & Edges  & Features & $\rho$  \\ 
\hline
Cora        & 7       & 2708  & 5429   & 1433              & 0.22              \\
CiteSeer    & 6       & 3327  & 4732   & 3703              & 0.36              \\
Pubmed & 3      & 19717 & 44338 & 50                & 0.52             \\
Wiki-CS     & 10      & 11701 & 216123 & 300               & 0.11              \\
\bottomrule
\end{tabular}
}}
\end{table}

\section{Experiments}\label{section5}
\subsection{Experiment Setup}\label{section5.1}
\subsubsection{Datasets and Evaluation Metrics}\label{section5.1.1}
We evaluate GraphALP on Pubmed \cite{2012Query}, Cora \cite{DBLP:conf/icml/LuG03}, CiteSeer \cite{DBLP:journals/aim/SenNBGGE08} and Wiki-CS \cite{mernyei2022wikics} datasets. Table \ref{table 1} shows the details of these datasets. For Wiki-CS, given the inherent class imbalance, we allocate 25\% of each class for training and validation while reserving 50\% for testing. For Cora, PubMed, and CiteSeer, the class distributions exhibit mild imbalance, we adopt a step imbalance setting by randomly selecting three classes as minority classes and down-sample them~\cite{DBLP:journals/nn/BudaMM18}.
Specifically, each majority class has 20 labeled nodes, whereas minority classes are scaled to $\rho \times 20$. Lower
$\rho$ denotes stronger imbalance.
In addition, we use Uniform Noise~\cite{DBLP:journals/tnn/SongKPSL23} as label noise, with probability $p$ of labels being randomly flipped to other classes.

We evaluate our model using accuracy (ACC), G-mean, and F1~\cite{DBLP:conf/www/LiuAQCFYH21}. While ACC effectively captures majority-class performance, it fails to reflect minority-class discrimination. Therefore, we employ F1 and G-mean to better assess balanced classification performance, where higher values indicate superior model capability.

\begin{table*}[ht]
\caption{ Ablation experiments on Cora and CiteSeer.}
\centering
\setlength{\tabcolsep}{6pt}
\begin{tabular}{ccccccccccccc}

\toprule
\multirow{2}{*}{Methods} & \multicolumn{3}{c}{Cora}                            & \multicolumn{3}{c}{CiteSeer}                        & \multicolumn{3}{c}{Pubmed}                          & \multicolumn{3}{c}{Wiki-CS}                         \\ \cline{2-13} 
                         & Acc             & G-Mean          & F1              & Acc             & G-Mean          & F1              & Acc             & G-Mean          & F1              & Acc             & G-Mean          & F1              \\ \hline
                
Ours       & \textbf{75.95} & \textbf{74.88} & \textbf{76.11} & \textbf{60.42} & \textbf{55.67} & \textbf{59.98} & \textbf{70.18} & \textbf{69.56} & \textbf{70.02} & \textbf{80.28} & \textbf{76.06} & \textbf{77.36} \\
Ours-PL    & 73.35          & 71.26          & 73.07          & 60.06          & 54.63          & 59.52          & 67.03          & 65.48          & 66.39          & 80.05          & 75.39          & 77.07          \\
Ours-RB    & 73.30          & 71.21          & 73.01          & 60.24          & 53.39          & 59.38          & 66.00          & 65.02          & 65.60          & 80.16          & 75.54          & 77.24          \\
Ours-W     & 73.35          & 70.39          & 72.67          & 59.52          & 53.26          & 58.79          & 67.64          & 66.14          & 67.09          & 80.19          & 75.85          & 77.30          \\
\bottomrule
\end{tabular}
\label{table 3}
\end{table*}

\subsubsection{Baselines and Parameter Settings}\label{section5.1.3}
We select two categories of baselines from class-imbalanced algorithms, focusing particularly on state-of-the-art approaches. (1) The representative approaches:  Vanilla~\cite{DBLP:conf/nips/HamiltonYL17}, Oversampling, Re-weight~\cite{DBLP:conf/ijcnn/YuanM12}, SMOTE~\cite{DBLP:journals/jair/ChawlaBHK02}, and Embed-SMOTE~\cite{DBLP:conf/pkdd/AndoH17}. (2) Data augmentation methods of class-imbalanced graphs: GraphMixup~\cite{DBLP:conf/pkdd/WuXGLTL22}, and GraphDAO~\cite{DBLP:journals/chinaf/XiaZZLY24}. Because our method focuses on mitigating the limitations of augmentation techniques on graphs.

The experiment is performed using PyTorch with the NVIDIA RTX A6000 GPU. All baselines are evaluated using the official parameters from GraphMixup~\cite{DBLP:conf/pkdd/WuXGLTL22}, except for GraphDAO which follows its original paper's settings. 
For our model, weight decay is $5e$-4, learning rate is $1e$-3. For LLM, we use DeepSeek-Chat\cite{DBLP:journals/corr/abs-2412-19437} to generate node text. For the LM, we adopt jina-embedding-v3\cite{DBLP:journals/corr/abs-2409-10173} to obtain node representations. 
The encoder layer and hidden dimension of AE and GAE are 2 and [64, 128], respectively. The decoder layer of GAE is 2. For Pubmed, Cora, and CiteSeer, the imbalance ratio is 0.7, the noise ratio is 0.3, and the oversampling scale is 0.8 unless otherwise specified. For Wiki-CS, we apply class-wise oversampling to balance classes. The results are obtained by averaging the outcomes of 5 runs.

\subsection{Overall Performance
}\label{section5.2}
The experimental results on node classification are shown in Table \ref{table 2}. 
We observed that (1) \textbf{Compared with traditional class imbalance methods, graph-based approaches demonstrate superior performance.} This is because conventional methods fail to account for relationships between nodes, resulting in incomplete information acquisition. In contrast, graph-oriented class imbalance techniques incorporate node dependencies, enabling the capture of richer structural information. (2) \textbf{GraphALP outperforms all baselines, achieving the highest performance.} For example, on Cora, GraphALP outperforms GraphDAO by 5.35$\%$ in ACC, 4.78$\%$ in F1, and 8.03$\%$ in the G-mean. GraphALP demonstrates a substantial lead in class-balance metrics F1 and G-mean. For instance, the model achieves significant G-mean improvements of 8.03\% (0.7488) on Cora and 6.03\% (0.7606) on Wiki-CS, evidencing its class bias mitigation capability. Moreover, the enhancement of ACC indicates that GraphALP strengthens the discrimination of minority classes without compromising overall accuracy. These results serve as a validation of the effectiveness of GraphALP.

\subsection{Robust Analysis}\label{section5.3}
To evaluate the robustness, we varied the class imbalance and label noise ratios on Cora, CiteSeer, and Wiki-CS.
\begin{figure}[ht]
\centering
\includegraphics[scale=0.525]{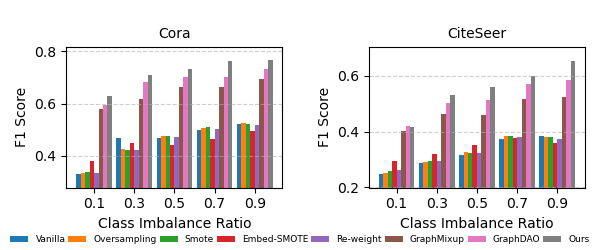}
\caption{Results under different class imbalance ratio.} 
\label{Fig. 4}
\end{figure}
\subsubsection{Influence of Imbalance Ratio}\label{section5.3.1}
To evaluate the robustness of GraphALP in relation to varying imbalance ratios, we maintain a fixed label noise ratio of 0.3 while varying the imbalance ratio across values of 0.1, 0.3, 0.5, 0.7 and 0.9. As Wiki-CS is an imbalanced class distribution naturally, imbalance ratio adjustment is not required. The results are illustrated in Figure \ref{Fig. 4}. We observed that (1) \textbf{As the imbalance ratio decreases, the performance of all methods deteriorates.} This observation indicates that the greater class imbalance exerts a more pronounced adverse effect, as the severe class imbalance induces biased representations of minority nodes toward the majority class, consequently compromising node classification accuracy. (2) \textbf{GraphALP is robust on different class imbalance conditions.} As the imbalance ratio increases, the node classification for GraphALP consistently outperforms the baselines. These results verify the robustness of GraphALP to varying imbalance ratios.   
\begin{figure}[ht]
\centering
\includegraphics[scale=0.525]{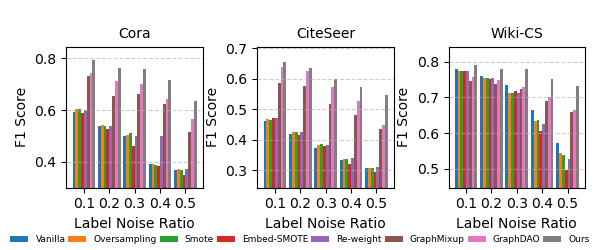}
\caption{Results under different label noise ratio.} 
\label{Fig. 5}
\end{figure}

\begin{figure*}[ht]

\centering
\subfigure[Vanilla]{
\includegraphics[width=0.16\linewidth]{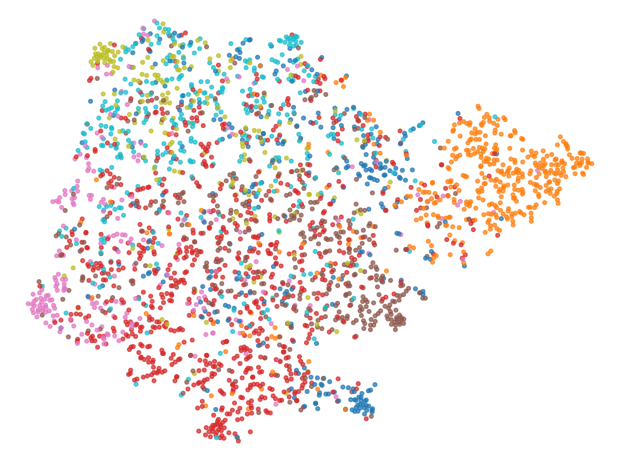}
}
\quad
\subfigure[Oversampling]{
\includegraphics[width=0.17\linewidth]{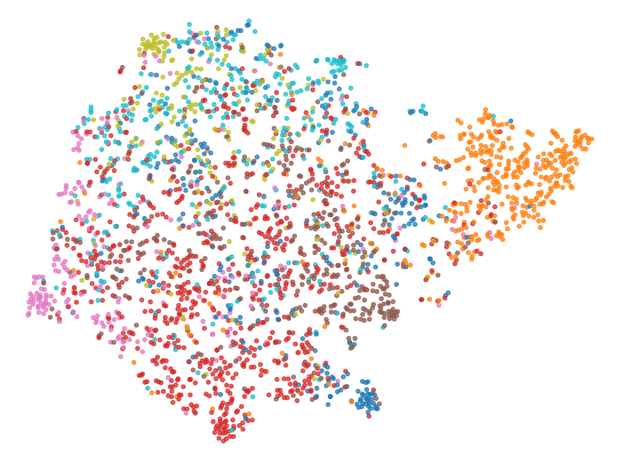}
}
\quad
\subfigure[GraphMixup]{
\includegraphics[width=0.17\linewidth]{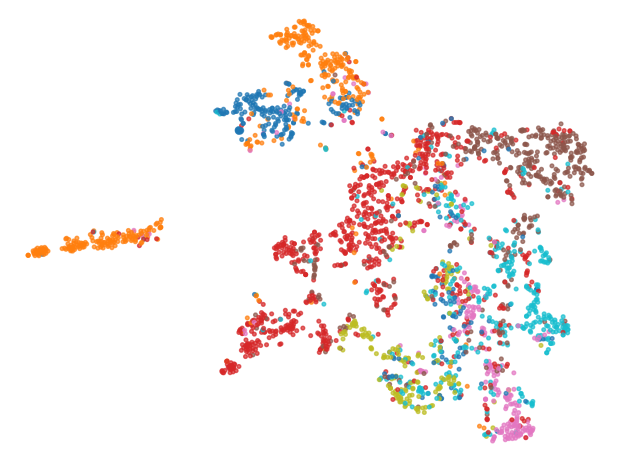}
}
\quad
\subfigure[GraphDAO]{
\includegraphics[width=0.17\linewidth]{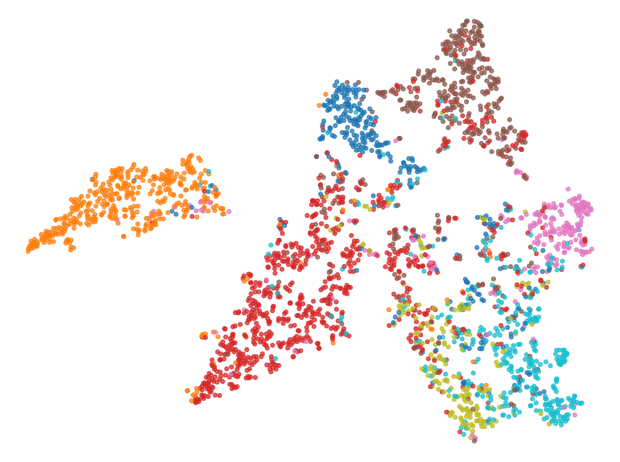}
}
\quad
\subfigure[Ours]{
\includegraphics[width=0.17\linewidth]{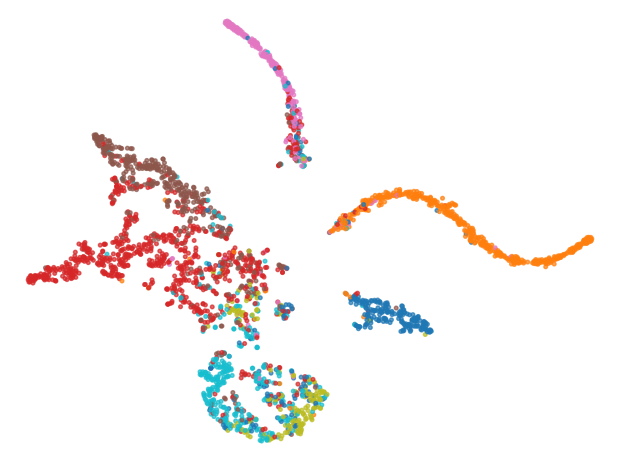}
}
\caption{Visualization of the Cora dataset.}
\label{Fig. 7}
\end{figure*}
\subsubsection{Influence of Label Noise Ratio}\label{section5.3.2}
To evaluate the robustness of GraphALP in relation to varying label noise ratios, we maintain a fixed imbalance ratio of 0.7 while varying the label noise ratio across values of 0.1, 0.2, 0.3, 0.4 and 0.5. The results are illustrated in Figure \ref{Fig. 5}. We observed that (1)
\textbf{As the label noise ratio decreases, the performance of all methods deteriorates.} This observation demonstrates that increased label noise exerts more significant detrimental effects, as higher noise ratios enable broader noise propagation through the GNN, ultimately degrading node classification accuracy. (2) \textbf{GraphALP is robust on different noise ratio.} As the noise ratio increases, the performance for GraphALP consistently outperforms the baselines. These results verify the robustness of GraphALP to varying noise ratios.
(3) \textbf{The improvement of GraphALP is more substantial when the extent of the noise ratio is more extreme.} For instance, in CiteSeer, when the label noise ratio is 0.1, GraphALP outperforms GraphDAO by 1.55\%, while the gap narrows to 9.78\% when the noise ratio reaches 0.5. 

\subsection{Ablation Study}\label{section5.4}
We conducted ablation studies to evaluate the contribution of each component by removing: (1) the pseudo-labeling module (Ours-PL), (2) the rebalancing module (Ours-RB), and (3) the weighted cross-entropy loss (Ours-W). These experiments are performed on PubMed, Wiki-CS, Cora, and CiteSeer. Table \ref{table 3} presents the ablation results, showing consistent performance drops when removing any module, confirming each component's contribution to model efficacy.
\subsection{Case Study}\label{section5.5}
To validate the efficacy of the pseudo-labeling approach and investigate the necessity of class rebalancing after pseudo-labeling, we conducted experiments by visualizing the training set with labels in the embedding space. 
Specifically, we inject 30\% uniform noise into Cora, generate node representations using our model, and conduct visualization analysis via t-SNE~\cite{DBLP:conf/aaai/TanLL00Z23}.
\begin{figure}[ht]
\centering
\includegraphics[scale=0.33]{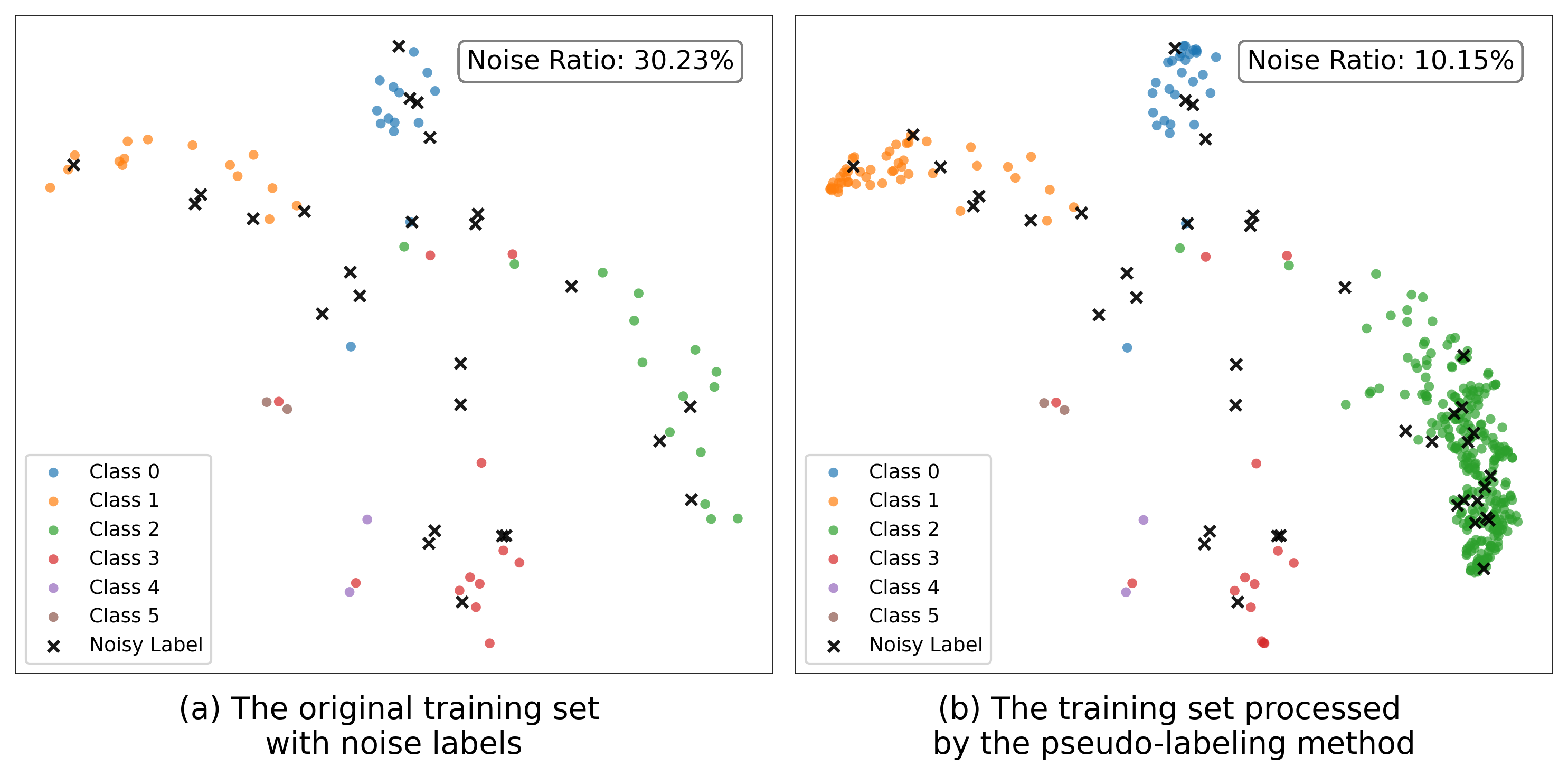}
\caption{Node representation visualization. Color dots and crosses are truth and noisy labeled nodes, respectively.} 
\label{Fig. 6}
\end{figure}

As shown in Figure \ref{Fig. 6}, GraphALP effectively reduces the label noise ratio. The results demonstrate that while the pseudo-labeling process introduces additional label noise, the noise ratio is reduced from 30.23\% to 10.15\%. Furthermore, as shown in Figure \ref{Fig. 6}(b), the pseudo-labeling process exacerbates the class imbalance, which experimentally justifies the essentiality of incorporating our LLM-based class rebalancing module following the pseudo-labeling module.

\subsection{Graph Visualization}\label{section5.6}
We employ t-SNE\cite{DBLP:conf/aaai/TanLL00Z23} to  visualize the node representations of GraphALP and baselines on Cora, enabling comparative analysis of the learned representations.
Specifically, We project the latent representations into 2D space using t-SNE with perplexity 30 and 5,000 iterations. Figure \ref{Fig. 7} shows the visualization results. The results demonstrate the superior performance of the GraphALP. Conventional class imbalance methods (e.g., Oversampling) exhibit poor clustering performance as they fail to incorporate graph structural information. Although advanced approaches (GraphMixup and GraphDAO) achieve significant improvements in clustering effectiveness, they still suffer from ambiguous class boundaries and frequent misclassification of minority nodes. In contrast, our proposed model not only maintains well-defined class boundaries but also achieves more compact intra-class node distributions.
\begin{table}[htbp]
\centering
\setlength{\tabcolsep}{1pt}
\caption{Comparison of parameter size and run time.}
\label{tab:param_time}

\begin{tabular}{l|c|c|c|c}
\hline
\textbf{Dataset} & \textbf{Architectures} & \textbf{GraphMixup} & \textbf{GraphDAO} & \textbf{Ours} \\
\cline{3-5}

\hline
Cora    & \makecell{Parameter (k)\\Time (s)}  & \makecell{487.5\\110.97}   & \makecell{1496.3\\120.39} & \makecell{1455.6\\96.53} \\\hline
CiteSeer & \makecell{Parameter (k)\\Time (s)}& \makecell{1068.4 \\151.03}  & \makecell{5264.4\\282.92} & \makecell{3641.5\\135.3} \\\hline
Wiki-CS  & \makecell{Parameter (k)\\Time (s)} & \makecell{197.3\\632.69}    & \makecell{712.7\\790.73}  & \makecell{364.7\\602.49} \\
\hline
\end{tabular}
\label{table 4}
\end{table}

\subsection{Practical Feasibility Analysis}\label{section5.8}	
Deploying graph learning methods in real-world scenarios presents two key challenges: (1) the co-existence of class imbalance and label noise necessitates a unified solution, and (2) naive method combination compromises effciency. To evaluate effciency, we summarize
the run time (training\&testing time) and parameter size in Table \ref{table 4}. We observe that GraphALP achieves lower runtime than both GraphMixup and GraphDAO, while maintaining a smaller parameter size than GraphDAO.
More importantly, as shown in Table \ref{table 2}, our model achieves significantly better performance.

\section{CONCLUSION}\label{section6}
In this paper, we propose a novel graph augmentation method with LLM and pseudo label (named GraphALP). GraphALP employs LLMs to generate synthetic minority nodes, effectively balancing the class distribution while reducing the label noise ratio. Furthermore, it incorporates a pseudo-labeling mechanism to expand the training set with high-confidence nodes, thereby enhancing supervisory signals and further decreasing the noise ratio. To address the class imbalance introduced by pseudo-labeling, we introduce the rebalancing module to mitigate the imbalance. These components operate synergistically, collectively addressing the dual challenges of class imbalance and label noise in graphs.  We conducted extensive experiments on node classification tasks, and the experimental results show that GraphALP obtains better performance than baselines.

\bibliography{aaai2026}


\end{document}